\documentclass[letterpaper]{article} 
\usepackage{include_aaai/aaai23}  
\usepackage{times}  
\usepackage{helvet}  
\usepackage{courier}  
\usepackage[hyphens]{url}  
\usepackage{graphicx} 
\usepackage{natbib}  
\usepackage{caption} 
\usepackage{algorithm}
\usepackage{algorithmic}

\usepackage{newfloat}
\usepackage{listings}
\DeclareCaptionStyle{ruled}{labelfont=normalfont,labelsep=colon,strut=off} 
\floatstyle{ruled}
\newfloat{listing}{tb}{lst}{}
\floatname{listing}{Listing}
\title{Go-Explore Complex 3D Game Environments for Automated Reachability Testing}
\author {
    Cong Lu\textsuperscript{\rm 1}\thanks{Work done at Microsoft Research, Cambridge, UK.},
    Raluca Georgescu\textsuperscript{\rm 2}\thanks{Joint senior authorship.},
    Johan Verwey\textsuperscript{\rm 2}\textsuperscript{\textdagger}
}
\affiliations {
    \textsuperscript{\rm 1} University of Oxford\\
    \textsuperscript{\rm 2} Microsoft Research\\
    cong.lu@stats.ox.ac.uk, raluca.georgescu@microsoft.com, joverwey@microsoft.com
}

\usepackage{xcolor}
\usepackage{amssymb}
\usepackage{amsmath}
\usepackage{cleveref}
\usepackage{booktabs}
\usepackage{multirow}
\usepackage{subcaption}

\begin{document}

\maketitle

\begin{abstract}
Modern AAA video games feature huge game levels and maps which are increasingly hard for level testers to cover exhaustively.
As a result, games often ship with catastrophic bugs such as the player falling through the floor or being stuck in walls.
We propose an approach specifically targeted at reachability bugs in simulated 3D environments based on the powerful exploration algorithm, Go-Explore, which saves unique checkpoints across the map and then identifies promising ones to explore from.
We show that when coupled with simple heuristics derived from the game's navigation mesh, Go-Explore finds challenging bugs and comprehensively explores complex environments \emph{without the need for human demonstration or knowledge of the game dynamics}.
Go-Explore vastly outperforms more complicated baselines including reinforcement learning with intrinsic curiosity in both covering the navigation mesh and number of unique positions across the map discovered.
Finally, due to our use of parallel agents, our algorithm can fully cover a vast 1.5km x 1.5km game world within 10 hours on a single machine making it extremely promising for continuous testing suites.
\end{abstract}

\section{Introduction}
With the scope and size of modern AAA games ever increasing, QA teams are struggling to exhaustively test these games~\citep{nantes2008framework, 8848091, doi:10.1177/1527476414525241}.
It is becoming increasingly common for games to ship with many bugs that are discovered by end-users and only get fixed over the course of several months following the title’s release~\citep{lewis2010went, 9402141}.
The challenge of covering every corner of a multi-square mile game world is exacerbated by the fact that the maps constantly change during development which requires tests to be repeated hundreds of times.

A large proportion of bugs that occur in game worlds are related to reachability~\citep{albaghajati2020video}.
Game testers need to verify that every area of a map that should be reachable through the player’s abilities, are in fact reachable and haven't been inadvertently blocked off during iteration.
Some areas are designed to test the player’s skill and are hard to reach, but not impossible.
Finally, there are those areas of the map that should not be reachable---in these areas, players could fall through the floor or find themselves inside geometry that is not part of the gameplay area. 
Automating reachability testing for large game maps would alleviate a huge burden during game development and allow bugs to be identified earlier without human labor.

\begin{figure}[t]
\centering
\includegraphics[width=0.99\columnwidth]{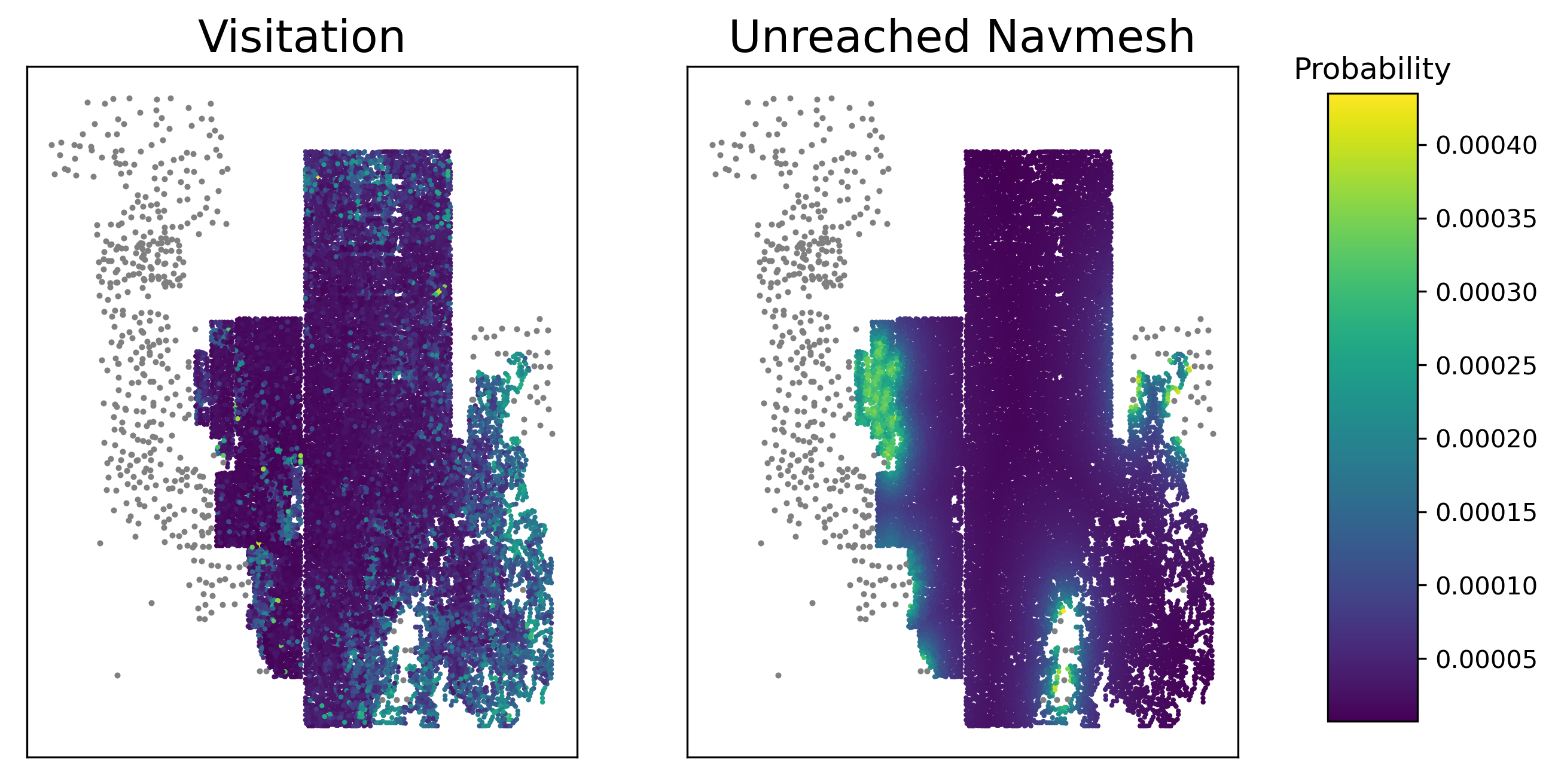}
\caption{
    Go-Explore caches discovered positions during training and then intelligently resets to promising saved positions to explore from them.
    The navigation mesh provides Go-Explore with a natural heuristic targeted towards unreached goals (marked in \textbf{\textcolor{gray}{gray}}) which may be combined with visitation count.
    Map shown is a top-down view of discovered positions on the hard traversal map (depicted in \Cref{subfig:traversal_pics}) 10\% through training.
    }
\vspace{-3mm}
\label{fig:go_exp_vis}
\end{figure}

A large body of prior work has considered ways to automate these tests including by imitating recorded human actions~\citep{8869824, 9463010, HERNANDEZBECARES201779}.
However, recorded actions are not robust to level changes, not exhaustive and cannot verify that forbidden areas are unreachable.
Other approaches include those that combine reinforcement learning with curiosity-based reward bonuses~\citep{ea1, ea2, Liu2022InspectorPA}.
Even so, these methods must use approximations for the vast observation spaces that feature in modern games, and have been shown under these settings to be brittle to catastrophic forgetting~\citep{goexplore1}.

\begin{figure*}[t!]
\centering
\begin{subfigure}{0.33\textwidth}
    \centering
    \includegraphics[width=.98\linewidth,height=.45\linewidth]{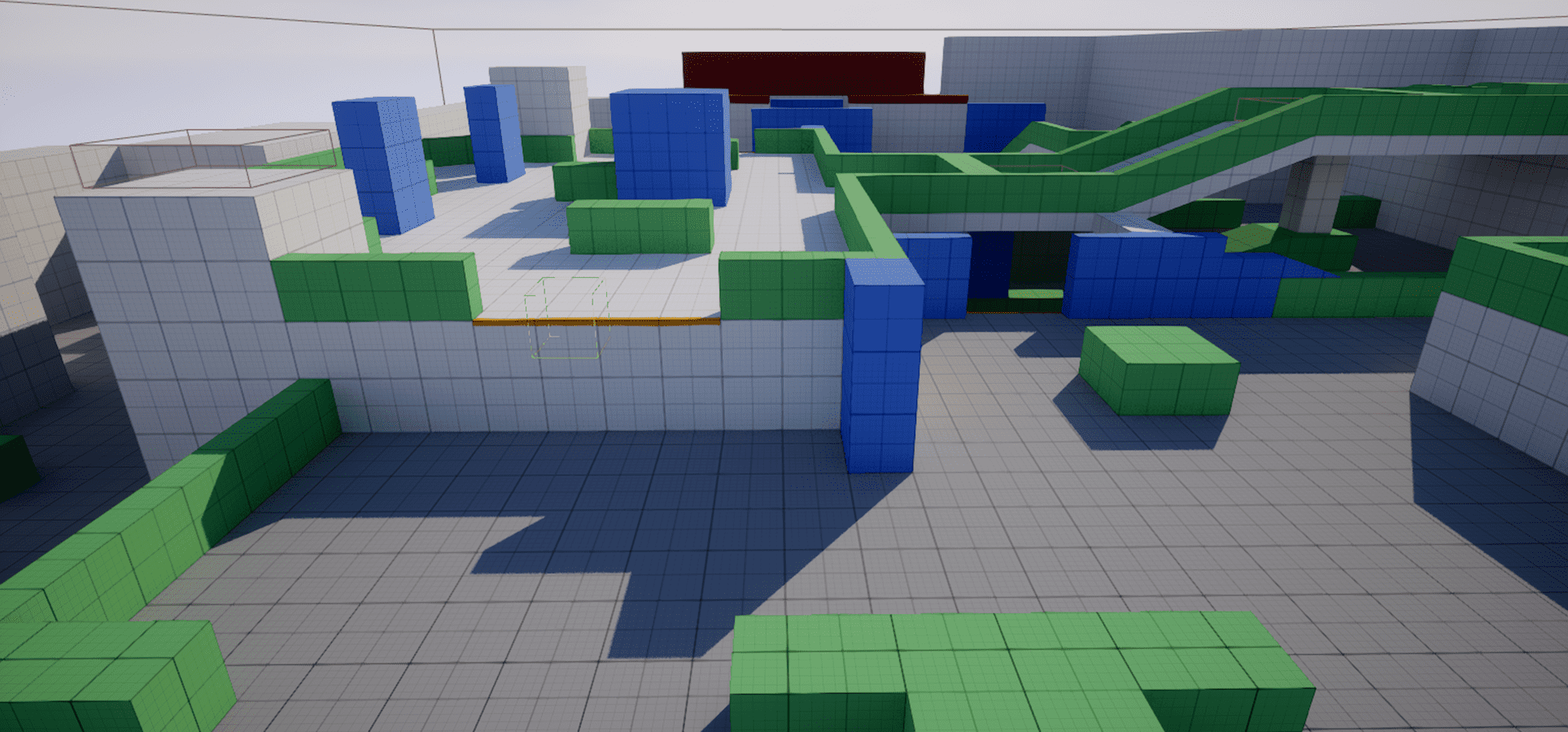}\\
    \vspace{1mm}
    \includegraphics[width=.98\linewidth]{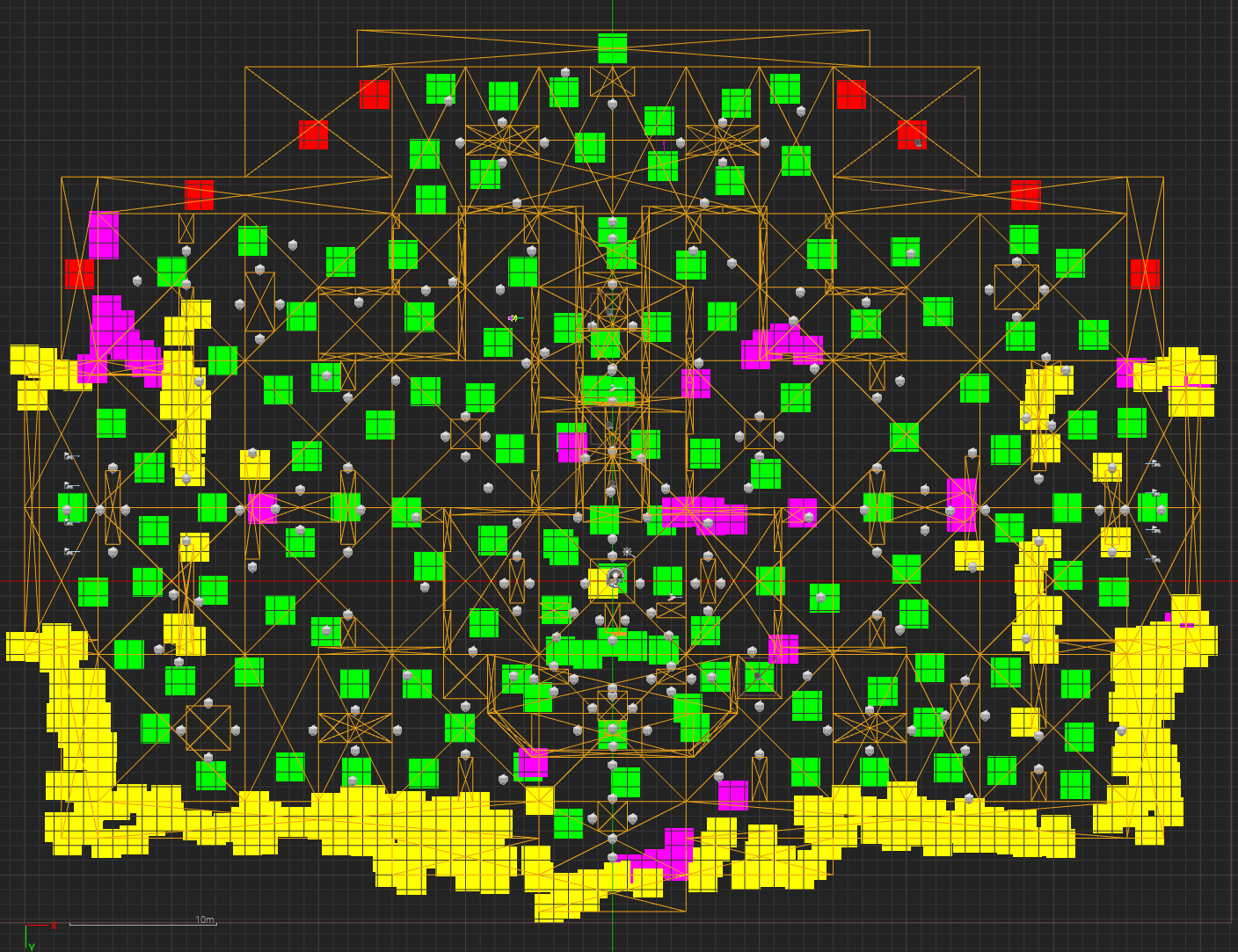}
    \caption{Small Map}
    \label{subfig:small_pics}
\end{subfigure}
\begin{subfigure}{0.33\textwidth}
    \centering
    \includegraphics[width=.98\linewidth,height=.45\linewidth]{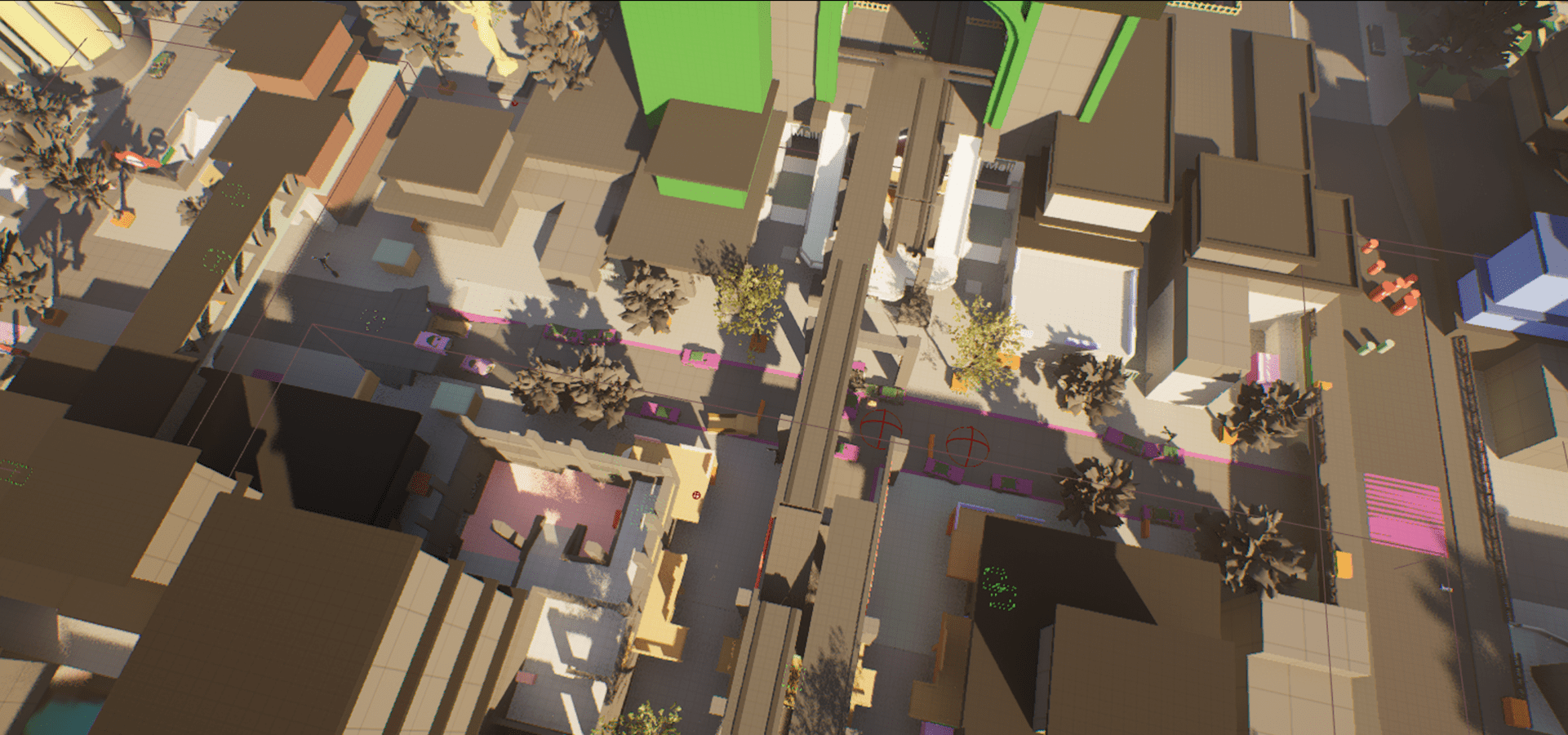}\\
    \vspace{1mm}
    \includegraphics[width=.98\linewidth]{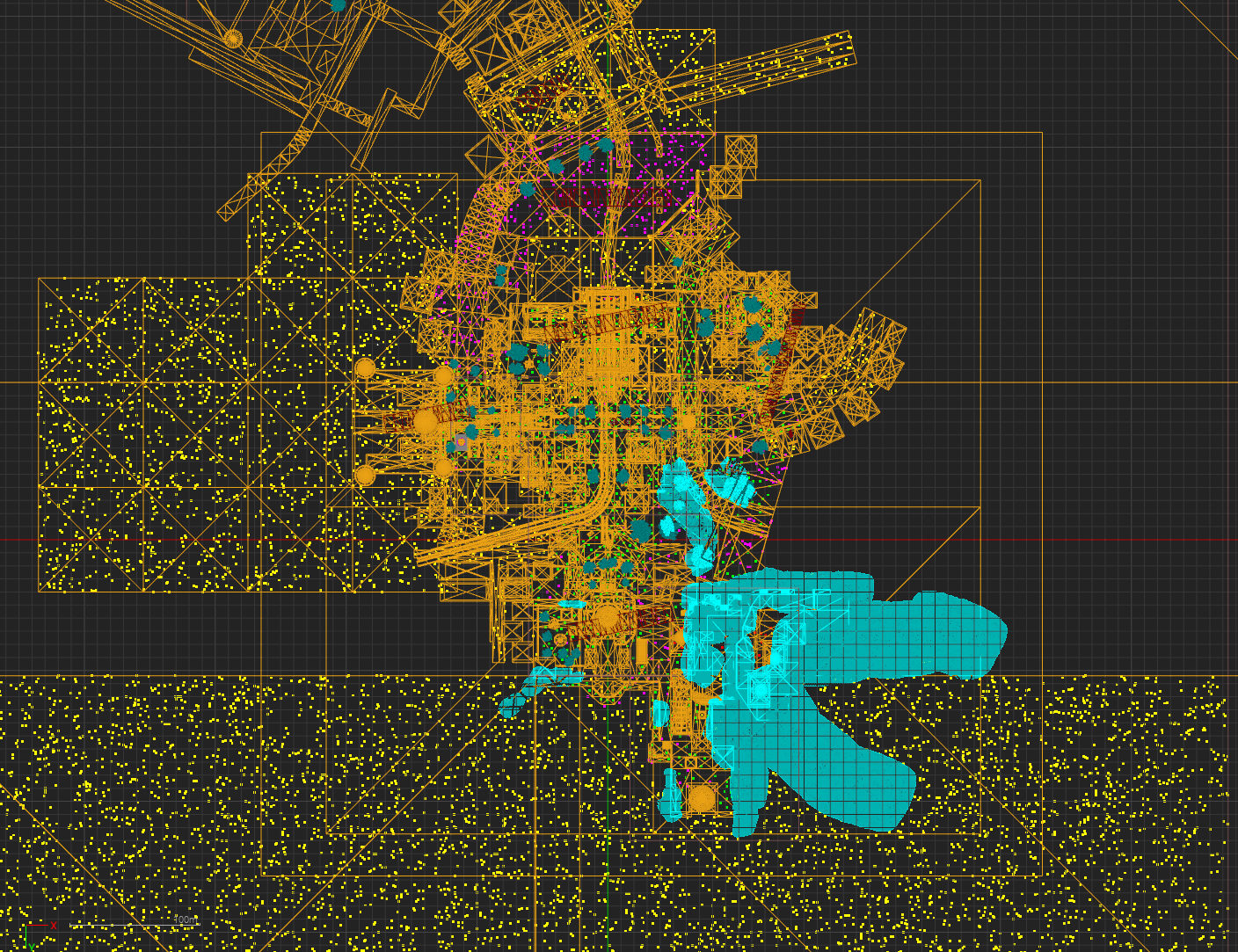}
    \caption{Large Map}
    \label{subfig:large_pics}
\end{subfigure}
\begin{subfigure}{0.33\textwidth}
    \centering
    \includegraphics[width=.98\linewidth,height=.45\linewidth]{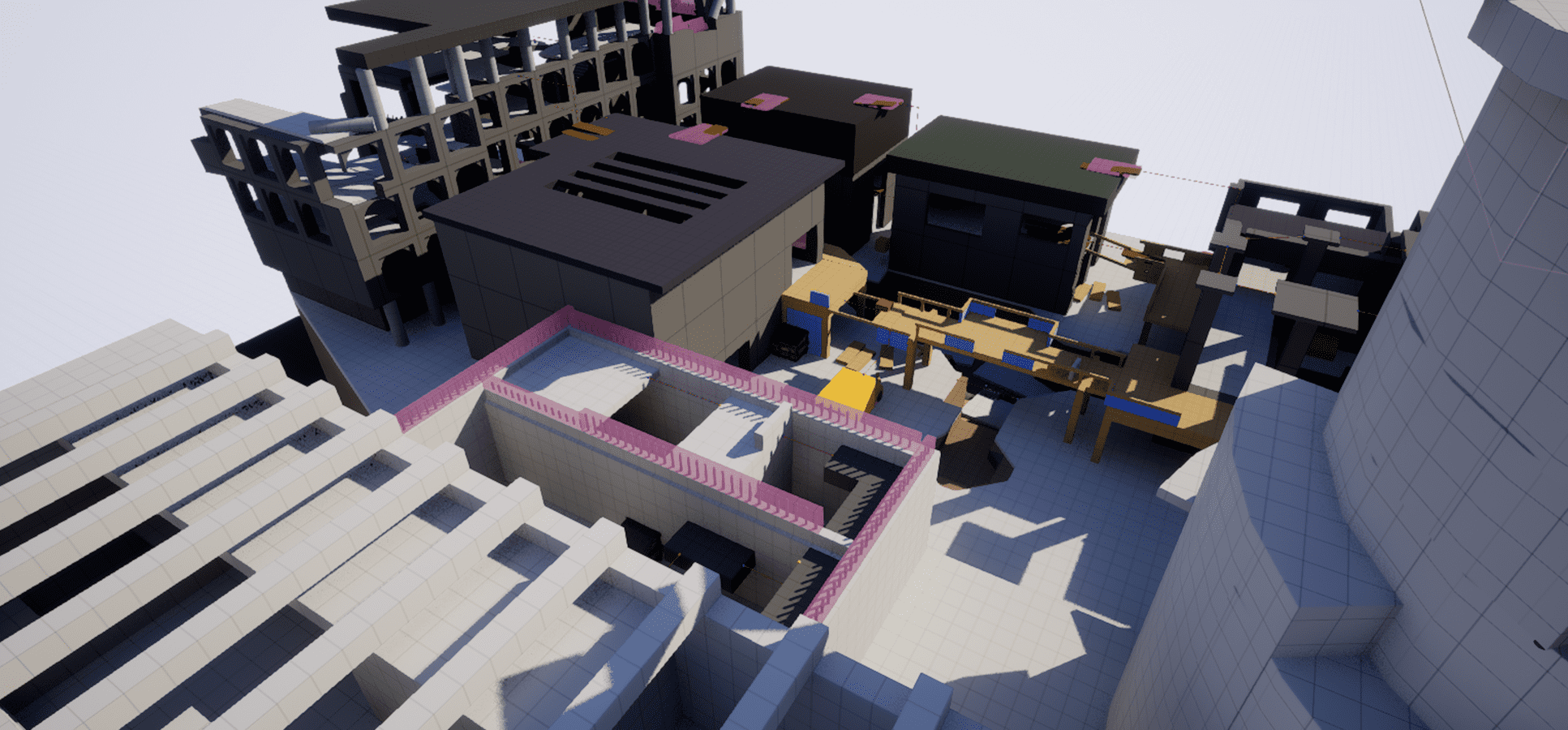}\\
    \vspace{1mm}
    \includegraphics[width=.98\linewidth]{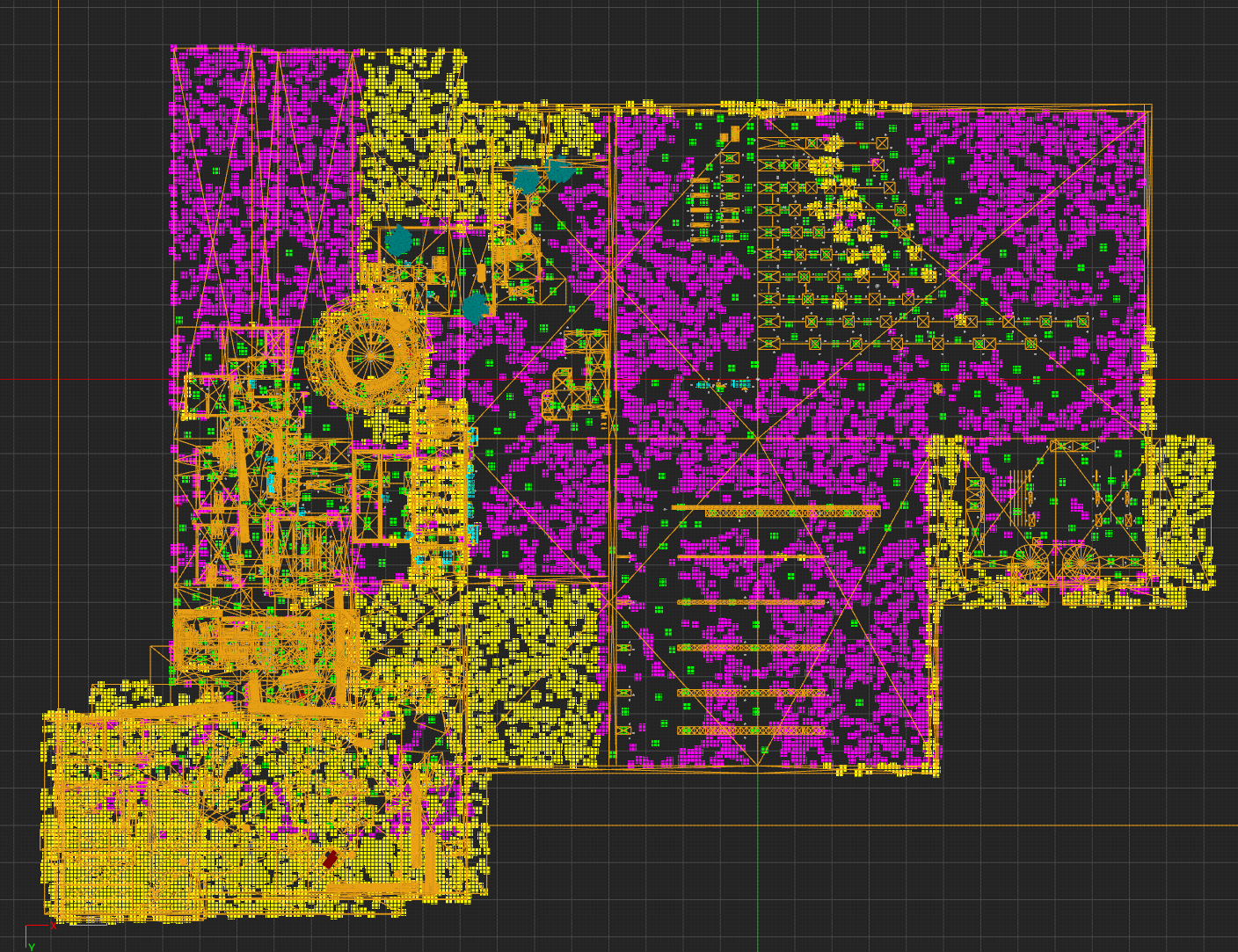}
    \caption{Traversal Map}
    \label{subfig:traversal_pics}
\end{subfigure}
\caption{
        We evaluate Go-Explore for reachability testing on three proprietary test game levels of increasing difficulty designed in Unreal Engine 5.
        The smallest is 100m x 100m and the largest 1.5km x 1.5km which matches the scale of modern game maps.
        We initially sample points from the nav-mesh which act as `goals' for exploration.
        Reached and unreached goals at the end of training are colored in \textbf{\textcolor{green}{green}} and \textbf{\textcolor{red}{red}} respectively.
        We additionally color non-goal points we discover as \textbf{\textcolor{magenta}{magenta}} if they are within 3m of the nav-mesh and \textbf{\textcolor{orange}{yellow}} otherwise.
        These \textbf{\textcolor{orange}{yellow}} points are discovered positions that an agent would not be able to reach by simply following the nav-mesh, and are often indicative of an unexpected bug.
        For example, on the small map, the \textbf{\textcolor{orange}{yellow}} points on the south wall point to an incorrectly configured section which allows the agent to reach the top of the wall and fall off the map leading to undefined behavior.
        \textbf{\textcolor{orange}{Yellow}} points on the left and bottom of the large map show places where the agent can escape the intended geometry.
    }
\vspace{-3mm}
\label{fig:pictures_of_envs}
\end{figure*}

In contrast, we propose to use Go-Explore~\citep{goexplore1}, a simple but powerful exploration algorithm which maintains a cache of discovered locations, and identifies promising locations with heuristics to reset and explore from.
We show that Go-Explore can readily handle parallelization and rapidly explore vast game worlds of many square miles.
Furthermore, we find empirically that random exploration is asymptotically better than learned exploration such as Random Network Distillation~\citep{rnd} under the same conditions which makes our approach easier to implement.

The core contributions of this paper are as follows:
\begin{enumerate}
\item We apply a simple and efficient algorithm, Go-Explore, for discovering reachable positions in hard exploration maps without human demonstrations or knowledge of the game dynamics.
\item We propose a criteria to classify discovered points into expected and unexpected by comparison to the game navigation mesh.
Undiscovered regions of the navigation mesh may also be flagged.
\item We show that by parallelizing agents within a single game instance, we can exhaustively cover vast game maps of the size 1.5km x 1.5km within 10 hours on a single machine, making our algorithm extremely promising for continuous testing suites.
\end{enumerate}

\section{Preliminaries}
\label{sec:prelims}
We begin by introducing traditional pathing in video games via navigation meshes, the reinforcement learning (RL) paradigm, and exploration algorithms in RL.
In this paper, we consider the problem of automatically testing for ``reachability'' which we define to be determining the set of locations of the map an agent can reach.
Note this is different from the traditional definition of graph reachability as we assume no knowledge of the environment dynamics to connect adjacent positions, and actions may have stochasticity.

\subsubsection{Navigation Meshes.}
AI agents in modern video games typically use a navigation mesh~\citep{navmesh, navmesh2002, graham2003pathfinding} or ``nav-mesh'' to navigate from one area of a map to another.
A nav-mesh is a collection of two-dimensional convex polygons~\citep{dunn11} which define areas of the map that are traversable by agents.
Within each polygon, an agent can freely move without being obstructed by any environment obstacles such as trees and rocks.
Adjacent polygons are connected together in a graph and pathfinding between them can be done with classic graph search algorithms such as A*~\citep{4082128}.

A na\"ive method of verifying reachability may be to send an agent to every node in the nav-mesh, one by one.
However, the nav-mesh may not cover the entire map and there are often gaps between patches of the nav-mesh that cannot be traversed by just walking.
Instead, we may use the nav-mesh as an initial starting point of positions we expect our agent to reach.
By comparing the results of an exploration algorithm versus this ground truth, we can \emph{categorize observed positions as expected or unexpected} and identify positions on the nav-mesh that we expect to reach but have not been reached which we illustrate in~\Cref{fig:pictures_of_envs}.

\subsubsection{Reinforcement Learning.}
A natural way to accelerate game testing would be to try and train agents to explore and find novel states.
Reinforcement Learning~\citep{kaelbling1996reinforcement, mnih-atari-2013,Sutton1998,tesauro1995temporal} is a successful paradigm for learning intelligent agents where optimal behavior may be specified by a reward function.
We model the game environment as a Markov Decision Process (MDP, \citet{Sutton1998}), defined as a tuple $M = (\mathcal{S}, \mathcal{A}, P, R, \rho_0, \gamma)$, where $\mathcal{S}$ and $\mathcal{A}$ denote the state space and action space respectively, $P(s' | s, a)$  the transition dynamics, $R(s, a)$  the reward function, $ \rho_0$  the initial state distribution, and $\gamma \in (0, 1)$ the discount factor.
The goal in RL is to optimize a policy $\pi (a | s)$ that maximizes the expected discounted return $\mathbb{E}_{\pi, P, \rho_0}\left[\sum_{t=0}^\infty \gamma^t R(s_t, a_t)\right]$.

\subsubsection{Exploration in RL.}
Count-based exploration describes a family of algorithms that assign reward bonuses based on visitation count to encourage agents to seek out novel states.
For finite state MDPs, we define the visitation count $n(s)$ of a state $s\in S$ to be the number of times a particular state has been encountered. 
Prior methods~\citep{pmlr-v70-ostrovski17a, bellemare_count, DBLP:conf/nips/TangHFSCDSTA17} have proposed to assign a reward bonus of $r_t = 1/n(s)$ or $r_t = 1/\sqrt{n(s)}$ at each time step.
These methods may be extended to MDPs with continuous state spaces by treating $n(s)$ as a density.
Even so, count-based methods are hard to scale to larger state spaces and cannot naturally handle environments in parallel.

Random Network Distillation (RND, \citet{rnd}) is a flexible way to compute an exploration bonus for high-dimensional state spaces by estimating the error of a neural network predicting features of the game observations given by a fixed randomly initialized neural network.
Concretely, the fixed randomly initialized \emph{target} network $f:S\rightarrow \mathbb{R}^k$ computes an embedding and the \emph{predictor} network $\hat{f}_\theta :S\rightarrow \mathbb{R}^k$ is trained by gradient descent to minimize the prediction error \mbox{$|| f(s) - \hat{f}_\theta(s) ||^2$} with respect to its parameters, $\theta$.
This prediction error is then used as the reward bonus.
Over the course of training the randomly initialized neural network $f$ is distilled into $\hat{f}_\theta$.
Intuitively, the prediction error is expected to be higher for novel states dissimilar to the ones that the predictor has been trained on.

\subsubsection{Go-Explore.}
Go-Explore~\citep{goexplore1, goexplore2} is an alternate approach for hard-exploration problems which maintains a cache of previously explored states and then uses heuristics to periodically select promising states and then explore from them.
Once sufficiently high return trajectories are found, the discovered behavior is made robust with imitation learning~\citep{hussein2017imitation}.
This algorithm aims to avoid the phenomenon of ``derailment'' that may occur in algorithms which do not have explicit memory such as RND where the predictor network may lose information about previously reached states due to catastrophic forgetting~\citep{doi:10.1073/pnas.1611835114, FRENCH1999128}.
This phenomenon could be even likelier to occur in large maps.

The first version of Go-Explore~\citep{goexplore1} assumes full access to a simulator of the environment and the ability to reset to any state previously seen, i.e. being able to choose $\rho_0$ throughout training.
This assumption is satisfied for the video games we test as we are able to save and restore simulator states.

\section{Go-Explore for Simulated 3D Environments}
\label{sec:reset}
In this section, we describe our adaptation of Go-Explore to 3D video games and the specific reset heuristics we use to identify promising previously discovered locations.
Since we do not need to follow the second phase of Go-Explore and imitate an optimal trajectory, the algorithm listed in~\Cref{alg:main_alg} simply builds a set of discovered positions $S_{\textrm{disc}}$ and thus resembles the first phase of Go-Explore with a tailored reset strategy.

Given a set of 3D positions $X$ and a distance metric $d$, we define the distance from the set $X$ to a point $y$, $d(X, y) := \min\{ d(x, y) | x\in X\}$.\footnote{This can be efficiently computed by data structures such as R*-Trees~\citep{beckmann1990r} which support nearest neighbor queries in amortized log time.}
We define two styles of reset heuristic: one based on state visitation counts, and the other guided by distance to the closest undiscovered point on the nav-mesh.

\begin{algorithm}[t]
\begin{footnotesize}
	    \caption{Compute Reachability (Exploration Phase of Go-Explore)}
	    \label{alg:main_alg}
	\begin{algorithmic}[1]
		\STATE \textbf{Input:} reset priority function $\eta:S\rightarrow \mathbb{R}$, total timesteps $T$, reset interval $t_{\textrm{reset}}$, distance threshold $K$
		\STATE \textbf{Initialize:} $S_{\textrm{disc}}=\emptyset$, set of visited positions
        \FOR{$t=1, \dots, T$}
        \STATE Observe state $s_t$
        \IF {$S_{\textrm{disc}}=\emptyset$ or $d(S_{\textrm{disc}}, s_t) > K$}
            \STATE $S_{\textrm{disc}}$.insert($s_t$)
        \ENDIF
        
        \IF {$t\mod t_{\textrm{reset}}$ = 0}
            \STATE Sample state $s_{t+1}$ from $S_{\textrm{disc}}$ using $\eta$
        \ELSE
            \STATE Take random action from $s_t$ to transition to $s_{t+1}$
        \ENDIF
    	\ENDFOR
	\end{algorithmic}
\end{footnotesize}
\end{algorithm}
\vspace{-3mm}

\subsubsection{Visitation-Based.}
We discretize the \mbox{$s=(x, y, z)$} positions in the game to a fixed granularity $K$ and then maintain visitation counts of each state.
This allows us to reset to a state in $S_{\textrm{disc}}$ with probability proportional to $\frac{1}{n(s)}$ where $n(s)$ is the visitation count.
By default, we measure distance in meters and use a granularity of $K=1$, i.e. a new position is deemed to be reached if it is more than one meter away from any previous position.

\begin{figure*}[t!]
\centering
\includegraphics[width=0.99\textwidth]{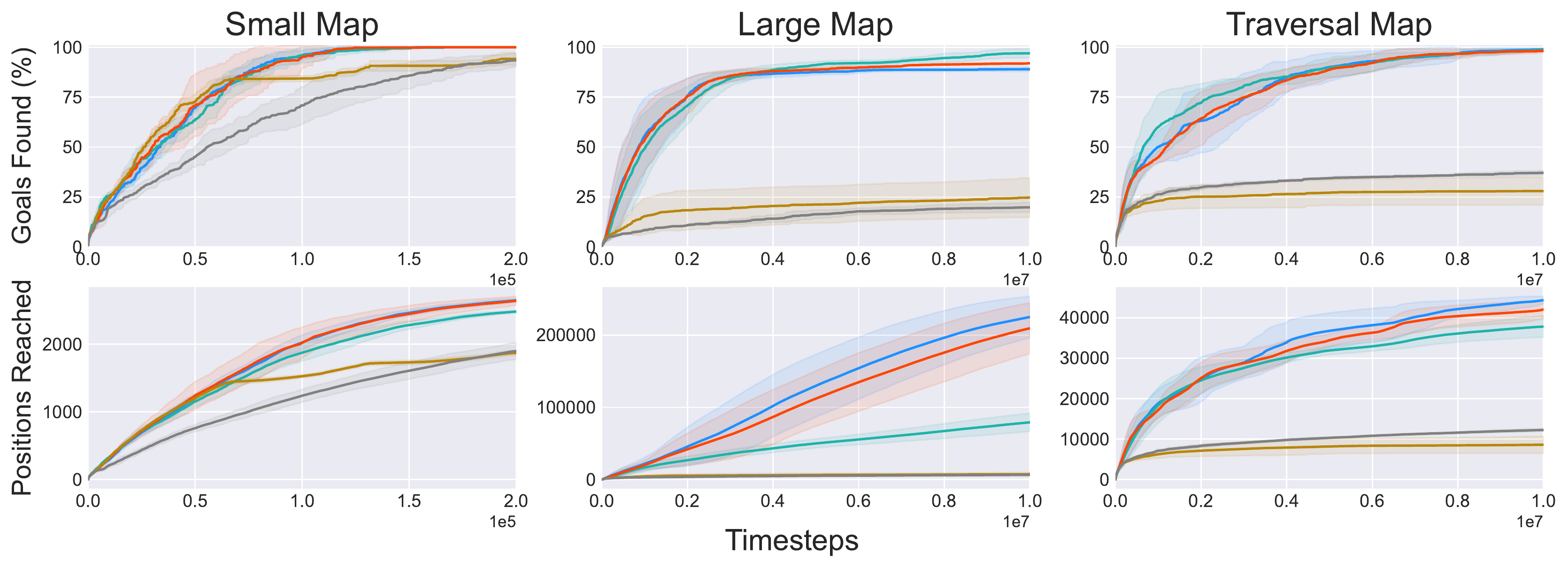}
\vspace{-1mm}
\includegraphics[width=0.85\textwidth, trim={0 100 0 100}, clip]{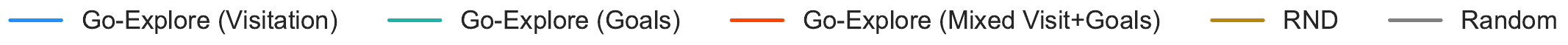}
\caption{
        Comparative evaluation of Go-Explore against the baselines showing the percentage of nav-mesh goals (illustrated in~\Cref{fig:pictures_of_envs}) reached on the top and unique positions discovered (up to 1m discretization) on the bottom.
        We note that Go-Explore vastly outperforms RND and pure random exploration especially on the larger maps and is reasonably stable to reset heuristic.
        Results show mean and standard deviation computed over 4 seeds.
    }
\vspace{-3mm}
\label{fig:main_eval}
\end{figure*}

\subsubsection{Nav-Mesh Goal Guided.}
We use the game nav-mesh to produce a set of points $S_{\textrm{nav-mesh}}$ or ``goals'' that we expect to be reachable but have not reached yet as a guide to exploration.
This set is constructed by sampling points at a pre-specified threshold at initialization.
Throughout training, we update $S_{\textrm{nav-mesh}}$ by draining any goals that are within one meter of a discovered position.
We may then reset to discovered states in $S_{\textrm{disc}}$ with probability proportional to $\frac{1}{d(S_{\textrm{nav-mesh}}, \cdot)}$.

We may also consider a weighted combination of the two reset heuristics as in~\citet{goexplore1} to define a custom reset priority function $\eta:S\rightarrow \mathbb{R}$.
This allows agents to balance between the two reset styles and avoid the failure mode of persistently attempting to reset close to an unreachable nav-mesh goal.
Once we reset to a promising state, we follow~\citet{goexplore1} and take random actions to explore.
We visualize the reset heuristics in~\Cref{fig:go_exp_vis}.
The notion of a goal generated from the nav-mesh could also be extended to goals on any interesting part of the map.

\section{Experimental Evaluation}
\label{sec:experiments}
We evaluate our approach on proprietary test game levels designed in Unreal Engine 5.
We consider three maps of increasing difficulty: 
\begin{itemize}
    \item \textbf{Small Map:} A small test bed with obstacles and two intentionally placed bugs (approx. 100m x 100m)
    \item \textbf{Large Map:} A vast cityscape to test the scalability of each algorithm (approx. 1.5km x 1.5km). This map is among the largest considered by modern automated approaches.
    \item \textbf{Traversal Map:} A large multi-level map with challenging geometry as a hard exploration challenge.
\end{itemize}
The game uses rich 3D observations with optional raycast, position and rotation features.
Agents move using a multi-discrete action space matching that of a standard game controller.
The maps are illustrated in~\Cref{fig:pictures_of_envs}.

For Go-Explore, we evaluate all reset heuristics previously described in the previous section---visitation-based, goal-based and mixed.
We use a default distance threshold of $K=1$ for adding a new point and use the same threshold for evaluating how many unique positions an agent visits over the course of training.
The agents are reset every $t_{\textrm{reset}} = 128$ steps.
We first evaluate the performance of Go-Explore against the baselines, RND (optimized using Proximal Policy Optimization, \citet{ppo}) and random exploration, across the three test maps.
Next, we ablate various components showing first, that Go-Explore is stable to hyperparameter choice and secondly, random exploration is critical for strong performance on large-scale maps.

For each experiment, we accelerate training by running 16 independent agents within each map that cannot interact with or see each other.
Each agent is synchronized with a central cache of discovered positions.
We train for 200K timesteps on the small map and 10M timesteps for both large maps, and repeat each experiment with 4 random seeds.
Due to our use of parallelization, wall-clock time for the small map is around 30 minutes, whereas the large maps only take 10 hours to cover on a single F8 Azure Virtual Machine.
We use a GeForce RTX 2080 GPU only for the RND baseline.

\subsection{Main Evaluation}
We first show that our algorithm can discover the intentionally placed bugs on the small map and that Go-Explore reliably outperforms the baselines across a variety of configurations.
Next, on the large maps, we find the difference between Go-Explore and the baselines grows more stark with the baselines making little progress.
On these maps, Go-Explore can cover the vast game worlds and discover many unexpected regions off-map.

\subsubsection{Bugs on the Small Map.}
\label{sec:toy}
\begin{figure}[t]
\centering
\includegraphics[width=0.99\columnwidth]{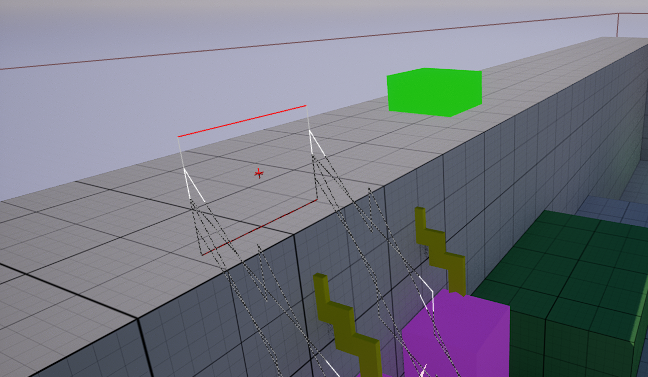}
\caption{
        Illustration of a bug on the small map surfaced by Go-Explore.
        The goal (in \textbf{\textcolor{green}{green}}) on the wall means the agent is unexpectedly able to reach the top of the wall and then fall off the map.
    }
\vspace{-3mm}
\label{fig:bug_small}
\end{figure}

We compare Go-Explore with all three reset heuristics on the left of~\Cref{fig:main_eval} evaluating coverage of the nav-mesh goals and unique positions found.
On this map, all three reset heuristics are comparable and are strong in terms of positions reached and goals found.
We manage to consistently find the two bugs that were placed on the small map, one where an incorrectly configured point led to agents being able to reach the top of the bounding wall and fall off the map which we illustrate in~\Cref{fig:bug_small} and another where an unreachable volume was incorrectly placed so an agent could incorrectly enter into a solid block.
The wall bug can also be identified in~\Cref{fig:pictures_of_envs} where the cluster of yellow points on the south wall of the map are unexpected and then can easily be flagged to the level developer.

\subsubsection{Large and Traversal Maps.}
\label{sec:large_maps}
We perform the same comparison as before on the larger maps as shown in the center and right side of~\Cref{fig:main_eval}.
On these maps, the difference between Go-Explore and the baselines grows more stark with Go-Explore being the only algorithm that can reliably cover at least the nav-mesh positions.
We also notice a difference between different styles of reset heuristic with Go-Explore.
On the large map, whilst the goal-based reset heuristic leads to the quickest coverage of the nav-mesh, it discovers far less unique positions than the reset heuristics based on visitation count.
This advocates for reset heuristics which can jointly encode information about undiscovered nav-mesh goals and visitation.

In~\Cref{fig:pictures_of_envs}, we can see discovered regions of both maps which are far away from the nav-mesh which are unexpected.
In the case of the traversal map, these correspond to being able to reach the top of large structures and fall off to unexpected regions.
This is a common source of reachability bugs and the automated flagging of problematic areas promises to save level designers and QA significant amounts of time.
We additionally note that Go-Explore scales well with map size---when comparing the small and large map, Go-Explore only takes $\sim$50x more timesteps to cover $\sim$200x more surface area, when comparing both nav-meshes.
Therefore, our algorithm may readily be integrated into nightly continuous testing suites as it \emph{only takes 10 hours on a single machine to fully cover a vast 1.5km x 1.5km game world}.
This could be further accelerated with parallel game instances.

Go-Explore is also efficient in terms of memory and time as we use a R*-Tree~\citep{beckmann1990r} to cache positions.
At the end of exploration on the large map, the $\sim$200K discovered points take around 1MB to store.
In a tree this size, checking and inserting a newly discovered point also takes less than a millisecond on average which is far less than the rate of simulation for the game.

\subsection{Ablation Studies}
In this subsection, we present ablation studies showing that Go-Explore is stable to hyperparameter choice; and showing the benefit of random exploration over RND even when both have smart resetting.
We further show ablations on Go-Explore using uniform sampling to show that the design of reset heuristic is important.

\begin{table}[t]
\centering
\caption{Ablations on the Small Map showing Go-Explore is robust to hyperparameter changes with all setups achieving 100\% of the goals before 200K timesteps. The default setting uses a threshold of 1m and a reset weight power of 1. Mean and standard deviation shown over 4 seeds.}
\label{tab:goexp_ablations}
\begin{tabular}{@{}lccc@{}}
\toprule
\multicolumn{1}{c}{\textbf{\begin{tabular}[c]{@{}c@{}}Reset\\ Type\end{tabular}}} &
  \textbf{\begin{tabular}[c]{@{}c@{}}Hyper-\\ parameters\end{tabular}} &
  \textbf{\begin{tabular}[c]{@{}c@{}}Timesteps to\\ All Goals\end{tabular}} &
  \textbf{\begin{tabular}[c]{@{}c@{}}Positions\\ Found\end{tabular}} \\ \midrule
\multirow{4}{*}{Goals}    & Default   & 128K         & 2485.2 $\pm$ 21.4          \\
                          & $K = 5$  & 110K         & 2564.0 $\pm$ \phantom{0}7.9           \\
                          & $p=1/2$ & 167K         & 2430.8 $\pm$ 37.6          \\
                          & $p = 2$   & 106K         & 2422.5 $\pm$ 14.8          \\ \midrule
\multirow{4}{*}{Position} & Default   & 119K         & 2649.5 $\pm$ 30.6          \\
                          & $K=5$    & \phantom{0}\textbf{90K} & \textbf{2736.0 $\pm$ 14.8} \\
                          & $p=1/2$ & 156K         & 2541.2 $\pm$ 21.3          \\
                          & $p = 2$   & 111K         & 2705.8 $\pm$ 36.8          \\ \bottomrule
\end{tabular}
\vspace{-3mm}
\end{table}

\subsubsection{Hyperparameters.}
We present ablations on the distance threshold $K$ required to add a new point to our set of visited positions $S_\textrm{disc}$ on the small map in \Cref{tab:goexp_ablations}.
We also include ablations on the power of the reset heuristic $p$, i.e. resetting using the heuristic $\eta^p$ instead of $\eta$.
The original Go-Explore paper used a power of $p=\frac{1}{2}$, i.e. taking the square root of all scores, however we find that $p=1$ works better for our use case which matches~\citet{10.1145/1553374.1553441}.
We see a marginal benefit from using $K=5$ for position based resetting, however we believe $K=1$ could lead to better checkpointing on hard exploration environments.

\subsubsection{RND vs. Random Exploration.}
We show that even when allowing the RND agent to reset to promising positions in the same way as Go-Explore on the large map, labeled ``RND + Go-Exp. Sampling'' in \Cref{fig:rnd_ablations}, the agent still under-performs Go-Explore (Random + Go-Exp. Sampling) in terms of final performance.
However, we do note that RND-based exploration (in dashed lines) does tend to be initially better but then trails off.
This could imply that the learned exploratory behavior from RND does not transfer to later regions of the map, perhaps preferring areas it has obtained reward from in the past.
This lends further support to our use of Go-Explore style \emph{random exploration} which requires no neural network overhead and is \emph{simpler to implement}.

\subsubsection{Go-Explore Heuristics vs. Uniform Sampling.}
\begin{figure}[t]
\centering
\includegraphics[width=\columnwidth]{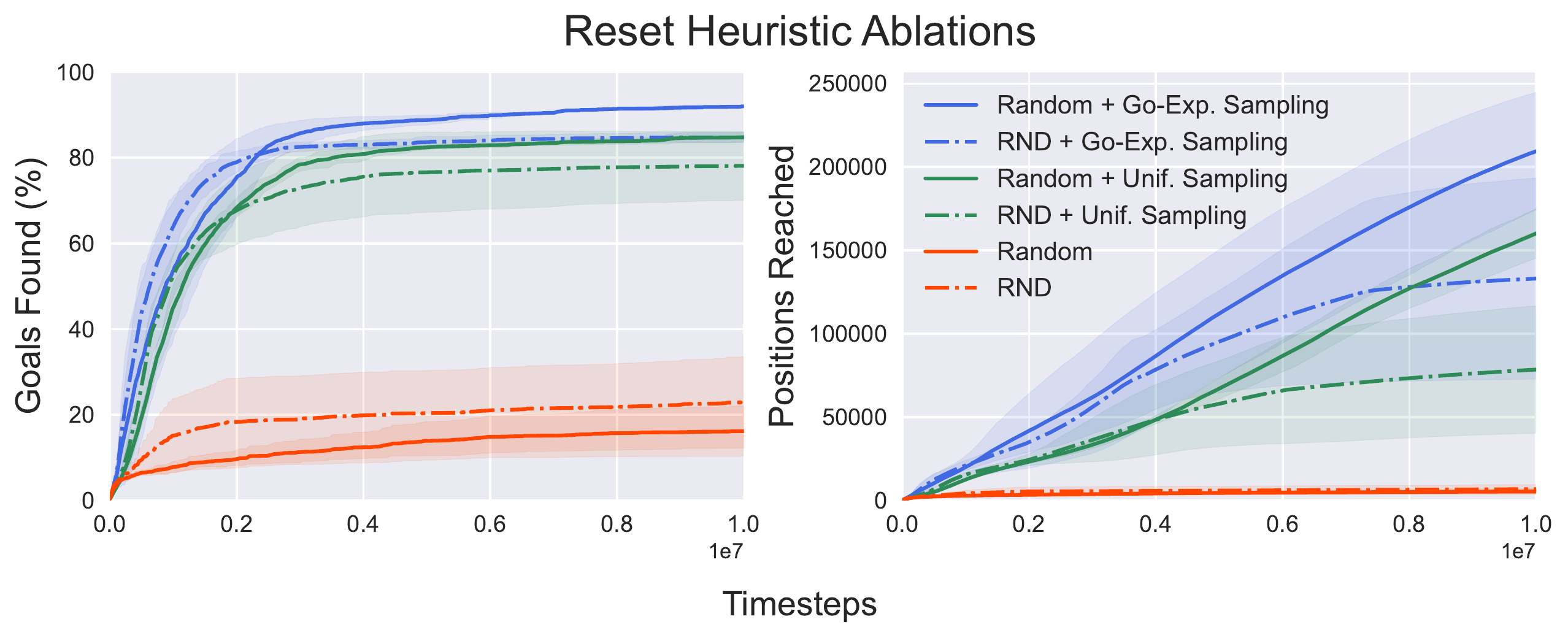}
\caption{
        Ablations for both random and RND-based exploration under visitation-based sampling (in \textbf{\textcolor{blue}{blue}}) and uniform sampling (in \textbf{\textcolor{green}{green}}) on the large map.
        This provides further support for purely random exploration, and shows that well-designed reset heuristics beat na\"ive uniform sampling.
        Results show mean and standard deviation computed over 4 seeds.
    }
\vspace{-3mm}
\label{fig:rnd_ablations}
\end{figure}

We further analyze the design choices involved in selecting sampling heuristics and ask how much better they are than just sampling visited points uniformly at random.
This corresponds to the comparison between Go-Exp. Sampling (in green) vs. Unif. Sampling in \Cref{fig:rnd_ablations} (in blue).
Although uniform sampling hugely boosts performance over simply resetting from the start position (in red), it does not reach the same levels of performance as with the carefully chosen heuristics.

\section{Related Work}
\label{sec:related_work}
Go-Explore has seen extensive use in discovering optimal policies for video games like Atari~\citep{goexplore1}; but to our knowledge, it has not seen use in reachability testing for large-scale modern game environments.

\subsubsection{Learning to Explore.}
Curiosity and imitation-based approaches where agents learn to explore have been successfully used to automate game testing in past work.
\citet{ea1} showed that a count-based exploration bonus with additional bonuses for reaching corners of the map approach with visitation-based resets was successful in covering a 500m x 500m map.
In contrast, we don't need a reward signal and our resets happen at a much higher frequency.
CCPT~\citep{ea2} used a hybrid approach combining RND and imitating human trajectories.
However, imitation-based approaches are liable to break with large map changes during development.
Inspector~\citep{Liu2022InspectorPA} combines RND with a separate module that attempts to identify and interact with key objects in a scene.
The interaction module could be a promising orthogonal extension to Go-Explore.
\citet{augmenting_bergdahl} consider an approach where an agent is reward on reaching pre-specified goals which inherently biases against exploring other parts of the game.

\subsubsection{Graph-based Search.}
Traditional search-based methods have shown promise in small games but are harder to scale.
Rapidly-Exploring Random Trees (RRT,~\citet{arxiv.1812.03125}) grows a state tree and labels edges with actions; however is limited by the size of the game action space.
CA-RRT~\citep{arxiv.2203.12774} improves on RRT by augmenting search with human demonstrations but is still only evaluated on relatively small maps.
SPTM~\citep{savinov2018semiparametric} navigates by building a graph of the environment and learning to retrieve nodes for planning.

\section{Conclusion}
In this paper, we show that Go-Explore allows for scalable, efficient and learning-free reachability testing of the vast 3D game maps typically found in modern AAA video games without the need for any human demonstration or knowledge of the game dynamics.
In contrast to prior curiosity-based RL approaches, Go-Explore avoids the issue of catastrophic forgetting by maintaining a cache of discovered positions.
The random exploration advocated by~\citet{goexplore1} is significantly simpler to implement and avoids the use of neural networks.
By flagging discovered points that are far away from the nav-mesh, we are able to cheaply identify reachability issues and highlight problematic points on the map.
Due to the fast run-time and ease of implementation of the algorithm, we expect Go-Explore may be readily integrated into a continuous testing framework.
In doing so, bugs may be found as each level is updated and significant resources may be saved in quality assurance.

\section*{Acknowledgments}
We would like to thank Sam Devlin, Anssi Kanervisto, Tabish Rashid, Tim Pearce, Mingfei Sun, and John Langford for reviewing earlier drafts of this work and many insightful discussions.
We are also grateful to the anonymous AIIDE 2022 Workshop on Experimental AI in Games reviewers for helpful and constructive feedback which helped to improve the paper.

\bibliography{references}

\end{document}